\documentclass[10pt,twocolumn,letterpaper]{article}

\usepackage{iccv}
\usepackage{times}
\usepackage{epsfig}
\usepackage{graphicx}
\usepackage{amsmath}
\usepackage{amssymb}

\usepackage{comment}
\usepackage{booktabs}
\usepackage{verbatim}
\usepackage{subcaption}
\usepackage{multirow}
\usepackage{algorithm}
\usepackage{algpseudocode}

\DeclareMathOperator*{\argmin}{\arg\!\min}

\usepackage[breaklinks=true,bookmarks=false]{hyperref}

\iccvfinalcopy 

\def\iccvPaperID{7146} 

\usepackage[capitalize]{cleveref}
\crefname{section}{Sec.}{Secs.}
\Crefname{section}{Section}{Sections}
\Crefname{table}{Table}{Tables}
\crefname{table}{Tab.}{Tabs.}

\usepackage{amssymb}
\usepackage{pifont}
\newcommand{\cmark}{\ding{51}}%
\begin{document}

\title{PNI : Industrial Anomaly Detection using Position and Neighborhood Information}

\author{Jaehyeok Bae$^{1,2}$ \ \ \ \ \ \ Jae-Han Lee$^1$ \ \ \ \ \ \ Seyun Kim$^1$ \\
{\large $^1$Gauss Labs Inc. \ \ \ \ \ \ $^2$Seoul National University} \\
{\tt\small wogur110@snu.ac.kr, \{jaehan.lee, seyun.kim\}@gausslabs.ai}
}

\maketitle

\begin{abstract}
Because anomalous samples cannot be used for training, many anomaly detection and localization methods use pre-trained networks and non-parametric modeling to estimate encoded feature distribution.
However, these methods neglect the impact of position and neighborhood information on the distribution of normal features.
To overcome this, we propose a new algorithm, \textbf{PNI}, which estimates the normal distribution using conditional probability given neighborhood features, modeled with a multi-layer perceptron network. Moreover, position information is utilized by creating a histogram of representative features at each position. Instead of simply resizing the anomaly map, the proposed method employs an additional refine network trained on synthetic anomaly images to better interpolate and account for the shape and edge of the input image.
We conducted experiments on the MVTec AD benchmark dataset and achieved state-of-the-art performance, with \textbf{99.56\%} and \textbf{98.98\%} AUROC scores in anomaly detection and localization, respectively.
\end{abstract}

\section{Introduction}
\begin{figure}
\centering
\subfloat{%
  \includegraphics[width=\columnwidth]{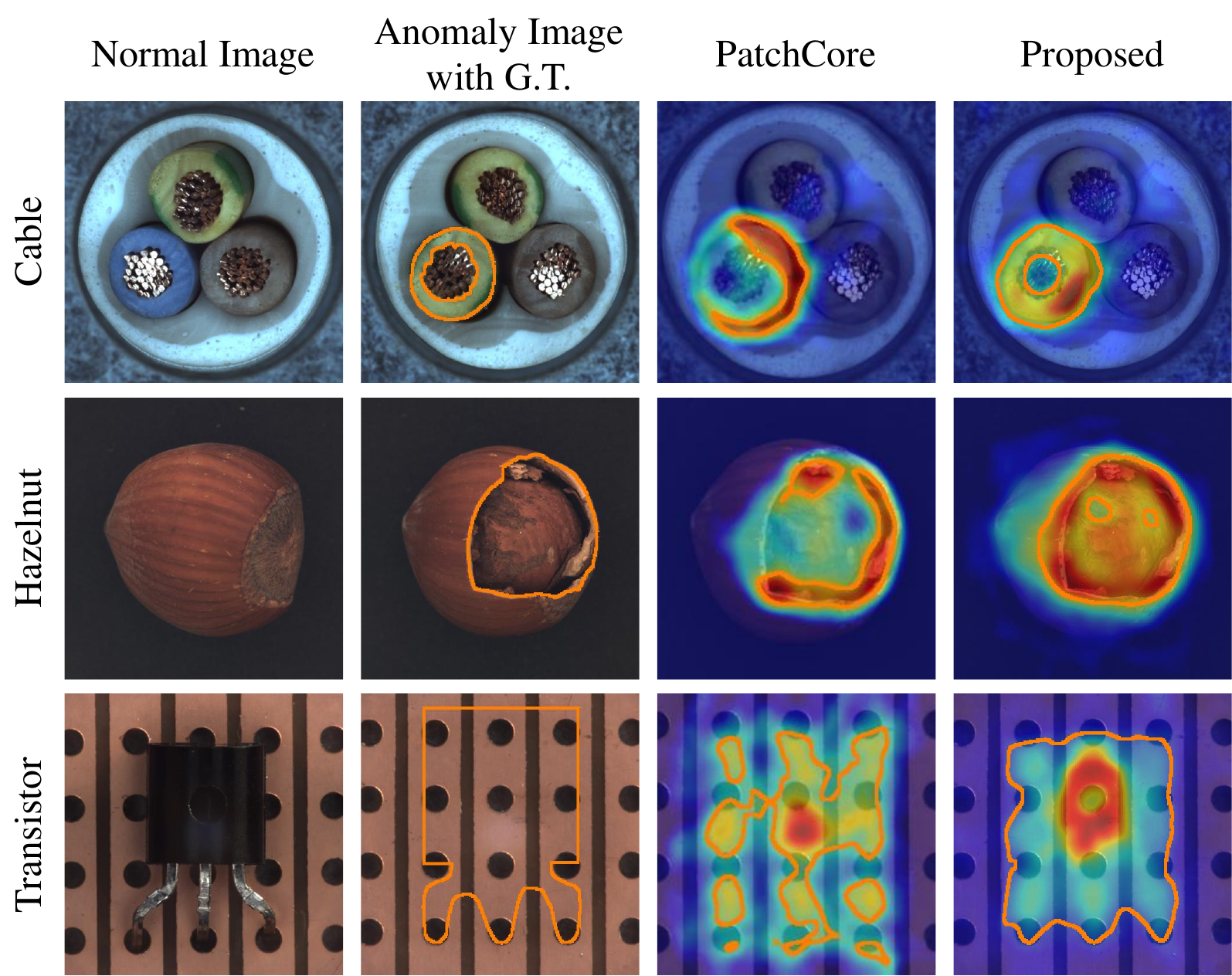}%
}
\caption{Examples from MVTec AD \cite{bergmann2019mvtec}.
The normal images (left) and the anomalous images (second column) overlaid by ground truth mask are followed by the anomaly maps from PatchCore \cite{roth2022towards} (third column) and our proposed approach (right).
The contours overlaid on anomaly maps are from thresholds optimizing F1 scores of anomaly localization.}
\label{Example_MVTec}
\vspace{-0.30cm}
\end{figure}

In industrial inspection \cite{bergmann2019mvtec}, delivering defective products to customers due to detection failure can be costly, and false detection can increase manufacturing costs.
Therefore, high prediction accuracy alone is insufficient, and low false positive rate (FPR) and false negative rate (FNR) are preferred.
Additionally, collecting abnormal samples can be difficult, making it almost impossible to build a supervised model for the task.
Thus, anomaly detection methods that use only normal samples are adopted. Anomaly localization quantifies the anomaly of each pixel in an input image, allowing users to identify where the defect is located and improve manufacturing processes.
Figure ~\ref{Example_MVTec} displays example images and results of an existing method and our proposed approach from the MVTec AD benchmark dataset.

Since there are only normal samples available for training, conventional classification methods cannot be used. One approach is to generate defective samples to train classifiers with supervised learning \cite{yi2020patch, yang2023memseg, schluter2022natural}. CutPaste \cite{li2021cutpaste}, for instance, used masks of rectangular and scar shapes to learn representation in a self-supervised manner. However, these methods show relatively low performance compared to other recent methods due to the lack of realism in abnormal patterns. To overcome this difficulty, many recently proposed methods \cite{defard2021padim, roth2022towards} adopt a pre-trained network such as pre-trained ResNet~\cite{he2016deep} using ImageNet~\cite{deng2009imagenet}, which intensively learned low-level image features from a super large-sized dataset.

Various methods are utilized to model the distribution of normal features that are transformed by a pre-trained network. For example, PatchCore \cite{roth2022towards} sub-samples representative features from the extracted normal features to achieve efficient non-parametric modeling of nominal features. Similarly, CFLOW-AD \cite{gudovskiy2022cflow} uses normalizing flow to model the normal feature distribution. However, these methods quantify the anomalies of input feature vectors independently, without considering the correlation between neighboring features. Additionally, normal features can be abnormal if they are in the wrong position. For instance, in the first row of Figure~\ref{Example_MVTec}, the top-view cables are normal only when the color order is correct, as shown in the first column image. However, if the color order is incorrect, the product is defective, even though all the local features are normal. Existing representation-based approaches, such as PatchCore, cannot capture this type of abnormality. Although CFLOW-AD adopted position encoding blocks, the implicit method appears insufficient to model positional information and overlooks the correlation between normal features.

To address this problem, this paper utilizes position and neighborhood information in simple yet effective ways. At each position of the encoded feature dimension, a histogram of all the training features is constructed to model a conditional probability distribution given the positional information. Meanwhile, an MLP (multi-layer perceptron) network models the probability distribution of normal features conditioned by neighboring features, where the input is the concatenated neighboring features. Through this process, the MLP network observes a large support region, while the features remain local, allowing the proposed method to produce a finely detailed localization map. These two distributions are combined to estimate the likelihood and anomaly score of an input image and its pixels during testing. While PatchCore serves as a baseline to demonstrate the validity of the proposed ideas, they can be applied to any representation-based method that uses a pre-trained network to generate input features and non-parametric modeling of normal features.

Representation-based approaches have a limitation in depicting detailed anomaly maps because local features are extracted from image patches of moderate size. When the patch size is small, enough information may not be extracted, leading to degraded detection performance. On the other hand, a large patch size may result in a blurred localization map. To overcome this problem, we trained an additional refinement network with synthetic abnormal images, which improves the detail of the localization map. It's important to note that synthetic images are not used to encode input images or to estimate anomaly scores. They are used only to train the refinement network. This is different from existing methods. By using synthetic images and corresponding anomaly maps generated by the above-mentioned method, the refinement network learns how to revise the anomaly map to look like the ground truth mask.

The proposed method resulted in a decrease in FNR from 1.83\% to 0.95\% compared to the current state-of-the-art method \cite{roth2022towards}. This reduction means that customers receive 48\% fewer defective products. Additionally, The FPR was reduced from 4.07\% to 1.50\%, which means that 63\% fewer good products are wasted.
Although the improvement in the area under the receiver operating characteristic (AUROC) metric, 0.46\%, may seem small, it can provide significant benefits to industrial manufacturing.

In summary, our contributions are threefold. Firstly, we demonstrate the effectiveness of using conditional normal feature distribution based on position and neighborhood information for anomaly detection and localization. Secondly, we validate that training a refinement network with synthetic datasets can significantly enhance performance. Finally, we provide insight into the factors that contribute to the noticeable improvement with the ablation study.
\section{Related Work}
\label{sec:related_work}
We selected PatchCore~\cite{roth2022towards} as our baseline because it employs a generic representation-based structure using non-parametric modeling and exhibits state-of-the-art performance. PatchCore aggregates local patch features from normal training data and selects a representative subset through greedy coreset subsampling~\cite{sener2017active}. During testing, the anomaly score for each patch feature is calculated pixel-wise by performing a nearest neighbor search from the coreset. Our proposed method utilizes the same process for feature vector creation, which we have summarized for completeness in this paper. However, it's worth noting that any other feature extraction process can be used because our proposed method is independent of the process used.

With a pre-trained network $\phi$, an input image is converted into hierarchical features $\phi_{i,j} = \phi_j(x_i)$, where $j$ denotes the hierarchy level of $\phi$. For instance, in ResNet-50 \cite{he2016deep}, $j \in {1,2,3,4}$ represents the final output of each resolution block. We denote the feature map $\phi_{i,j} \in \mathbb{R}^{c^j \times h^j \times w^j}$ as a three-dimensional tensor, where $c^j$, $h^j$, and $w^j$ are the number of channels, height, and width, respectively.
To avoid using too high or low-level features, the intermediate features $\phi_{i,2}$ and $\phi_{i,3}$ are concatenated and used. As the spatial sizes $(h,w)$ of these features are different, the smaller one is resized to be the same size as the larger one: $(h^*, w^*)$, where $h^*=\max(h_2,h_3)$ and $w^*=\max(w_2,w_3)$. Then, they are concatenated to obtain $\phi^*_i \in \mathbb{R}^{c^* \times h^* \times w^*}$, where $c^* = c_2 + c_3$. Furthermore, to increase the receptive field of feature maps, the pixel-level feature $\Phi_i(h,w)$ is extended to incorporate neighborhood features within a specific patch size $p$.
Adaptive average pooling is performed to output a single feature of dimension $d$ at $(h,w)$. Through this process, the input image is converted into a set of local patch-level features $\Phi_i \in \mathbb{R}^{d \times h^* \times w^*}$, where $d$ denotes the dimension of the feature vector.

\begin{figure*}[t]
\centering
\includegraphics[width=\linewidth]{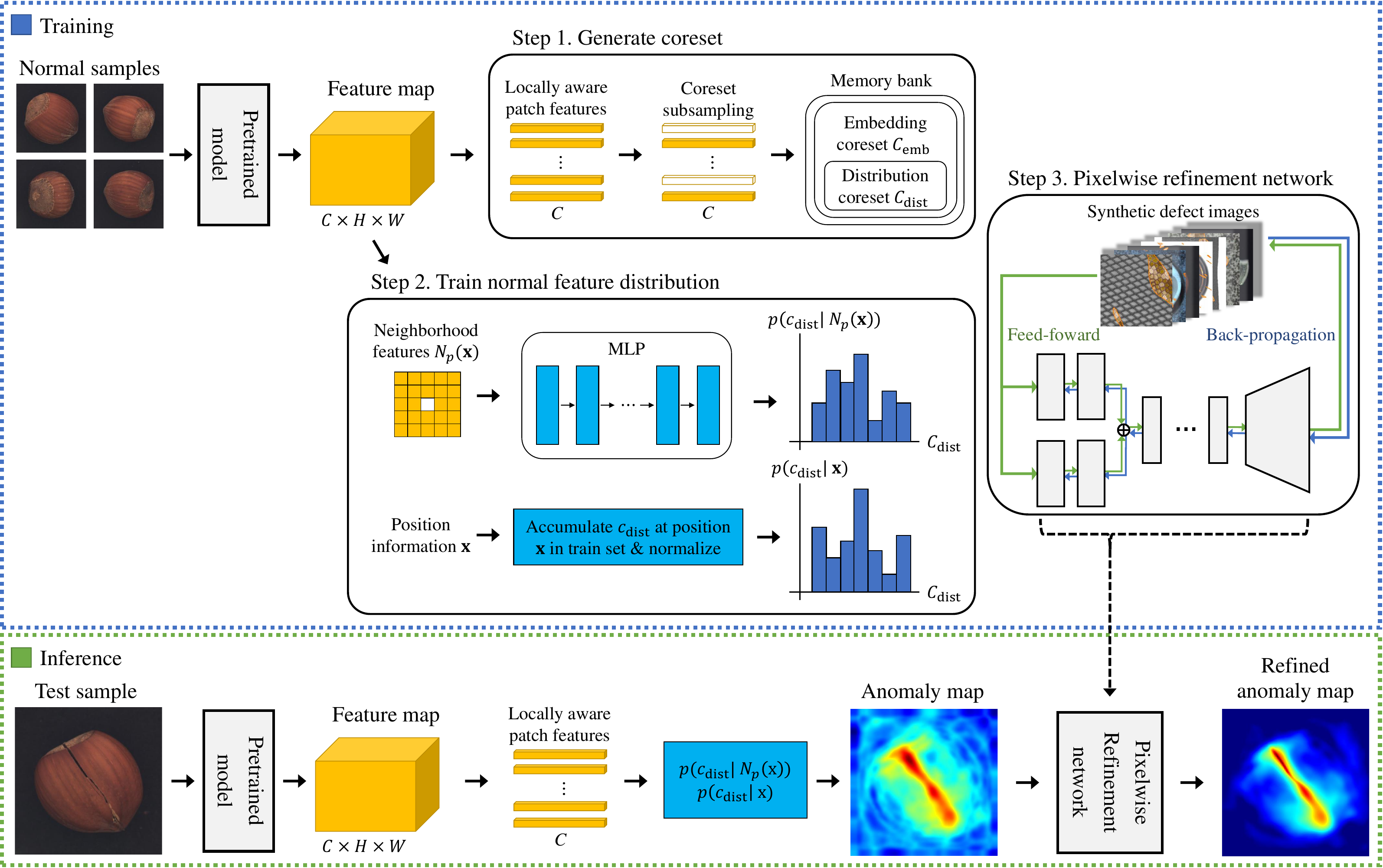}
\caption{Overview of PNI algorithm. At train time, normal samples are converted to feature map $\Phi_i$ using ImageNet pre-trained model $\phi$. Aggregated patch-level features are subsampled to generate embedding coreset $C_\mathrm{emb}$ and distribution coreset $C_\mathrm{dist}$ using the greedy subsampling method. After storing the coresets, normal feature distribution given neighborhood and position information is trained with MLP and histogram respectively. A pixel-wise refinement network is trained separately using synthetic defect images. At inference time, anomaly score for local test feature is evaluated using the trained normal feature models. At last, the refinement step is performed to improve the anomaly map considering the input image.}
\label{Overview}
\vspace{-0.30cm}
\end{figure*}

\section{Method}
\subsection{Overview}

Assume that $\mathbf{x}$ represents the spatial coordinates $(h,w)$ in the patch-level feature $\Phi_i$.
In most existing representation-based methods \cite{cohen2020sub, defard2021padim, roth2022towards}, the anomaly score of the patch-level feature $S(\mathbf{x})$ is estimated as the negative log-likelihood of $p(\Phi_i(\mathbf{x}))$, given by
\begin{equation}
\label{negative log likelihood}
S(\mathbf{x}) = -\log p(\Phi_i(\mathbf{x})),
\end{equation}
where $p(\Phi_i(\mathbf{x}))$ represents the probability that $\Phi_i(\mathbf{x})$ is normal and is modeled using trained normal features.

In this paper, we argue that the probability should be modeled based on the position and neighboring features.
As shown in Figure \ref{Example_MVTec}, electric wires are located within their sheaths (as seen in the left image of the first row), while transistors in the normal dataset are typically located at the center of the images (as seen in the left image of the third row).
Denoting position and neighborhood information as $\Omega$, the anomaly score $S(\mathbf{x})$ is represented as negative log-likelihood of conditional probability of $\Phi_i(\mathbf{x})$ given $\Omega$:
\begin{equation}
\label{nll with omega}
S(\mathbf{x}) = -\log p(\Phi_i(\mathbf{x}) | \Omega).
\end{equation}

To model the conditional probability from training features, we introduce the embedding coreset $C_\mathrm{emb}$.
The feature vectors of $C_\mathrm{emb}$ are sub-sampled from all normal features in all training images using a greedy coreset sub-sampling method \cite{sener2017active}.
Each element of $C_\mathrm{emb}$ delegates a group of similar normal features.
In a given circumstance, the number of occurrences of normal features associated with an embedding coreset vector $c \in C_\mathrm{emb}$ is proportional to the probability that $c$ is normal in that condition $p(c | \Omega)$.
The normal probability of a patch $p(\Phi_i(\mathbf{x}) | \Omega)$ is expressed with $C_\mathrm{emb}$ as follows:
\begin{equation}
\begin{aligned}
\label{likelihood0}
p\left(\Phi_i(\mathbf{x}) | \Omega \right) = \sum_{c \in C_\mathrm{emb}} p(\Phi_i(\mathbf{x}) | c, \Omega) \ p(c | \Omega).
\end{aligned}
\end{equation}
While the computation is challenging with a large size of $C_\mathrm{emb}$, it has been observed that $p(c | \Omega)$ is a sparse distribution with many small values that can be ignored.
To take advantage of this property, \eqref{likelihood0} is approximated as follows:
\begin{equation}
\begin{aligned}
\label{final_likelihood}
p\left(\Phi_i(\mathbf{x}) | \Omega \right) \approx \max_{c \in C_\mathrm{emb}} p\left(\Phi_i(\mathbf{x}) | c, \Omega \right) \ T_{\tau}\left(p(c | \Omega)\right),
\end{aligned}
\end{equation}
where $T_{\tau}(x)$ is defined as:
\begin{equation}
\begin{aligned}
\label{threshold}
T_{\tau}(x) =
\begin{cases}
1,\; \mathrm{if}\;x >\tau \\
0,\; \mathrm{otherwise}.
\end{cases}
\end{aligned}
\end{equation}
To stop considering insignificant values of $p(c | \Omega)$, the threshold function is applied.
Applying $T_{\tau}(x)$, $p(c | \Omega)$ with moderate probability becomes one.
Using this thresholding technique and the maximum operation in equation \eqref{final_likelihood}, it is possible to identify the coreset feature that is most similar to $\Phi_i(\mathbf{x})$ while rejecting improbable $c_\mathrm{emb}$ features.
While this approximation may not be intuitive, it can significantly reduce computation time with only a small decrease in performance.
$\tau$ lower than $1 / |C_\mathrm{emb}|$ guarantees at least one of $c$ in $C_\mathrm{emb}$ be a normal feature.
In this paper, we set $\tau = 1 / (2|C_\mathrm{emb}|)$ without optimizing.

To generate an anomaly score map, scores are estimated for all features in an input image. However, the resolution of the score map may differ from that of the original input, so it is resized using bi-linear interpolation and smoothed with a Gaussian kernel of $\sigma = 8$ as described in \cite{roth2022towards}. Note that the parameter $\sigma$ is not extensively optimized. While Gaussian smoothing is performed to eliminate noisy values, it may damage the detailed information of the score map. Therefore, an additional pixel-wise refinement step is performed to enhance the resized score map and make it more consistent with the edges, textures, and shapes of defects and objects in the input image.

\subsection{Modeling Normal Feature Distribution}
To model the normal feature distribution $p(c|\Omega)$, it is approximated as an average of the two probabilities as:
\begin{equation}
\begin{aligned}
\label{prob_pni}
p(c|\Omega) & \approx \frac{p(c|N_p(\mathbf{x})) + p(c|\mathbf{x})}{2},
\end{aligned}
\end{equation}
where $p(c|N_p(\mathbf{x}))$ is the normal feature distribution in neighborhood information and it is modeled using an MLP.
To model $p(c|\mathbf{x})$, the normal feature distribution in position information, histograms are constructed by counting the normal training features at every position $\mathbf{x}$ as shown in Figure \ref{Overview}.
To train the MLP and create the histograms, using a small size for $C_\mathrm{emb}$ is preferable.
However, reducing the size of $C_\mathrm{emb}$ can lead to a decrease in the accuracy of the normal probability of the input vector, $p(\Phi_i(\mathbf{x}) | c, \Omega)$.
To address this issue, we introduce the distribution coreset, $C_\mathrm{dist}$, which is sub-sampled from the embedding coreset $C_\mathrm{emb}$ using the same method as in \cite{sener2017active}.
In our implementation, both coresets are calculated simultaneously because $C_\mathrm{dist}$ is a subset of $C_\mathrm{emb}$, and the mapping from $c_\mathrm{emb}$ vectors to $c_\mathrm{dist}$ vectors is calculated at the beginning.
Therefore, $p(c_\mathrm{emb}|\Omega)$ in equation \eqref{final_likelihood} is changed to $p(c_\mathrm{dist}|\Omega)$ according to the corresponding $c_\mathrm{dist}$ vectors.

To model $p(c_\mathrm{dist}|\Omega)$, a simple MLP network is trained with neighboring features $N_p(\mathbf{x})$ of input feature $\Phi_i(\mathbf{x})$.
$N_p(\mathbf{x})$ is defined as a set of features that are within a $p \times p$ patch, excluding $\mathbf{x}$ itself as follows:
\begin{equation}
\begin{aligned}
\label{neighborhood feature generation}
N_p(\mathbf{x}) = \{\Phi_i(m, n) \mid |m-h| \leq p/2, \hspace{1.8cm}\\
\hspace{1.8cm} |n-w| \leq p/2, (m, n) \neq (h,w) \},
\end{aligned}
\end{equation}
where $\Phi_i(m, n)$ is feature vector at position $(m, n)$ in the feature map $\Phi_i$.
The MLP takes an input of a 1-dimensional vector obtained by concatenating all features in $N_p(\mathbf{x})$ and has $N_\mathrm{MLP}$ sequential layers with $c_\mathrm{MLP}$ channels.
Batch normalization and ReLU activation functions are used between layers.
The output of the MLP has $|C_\mathrm{dist}|$ nodes, with the value of each node representing the probability of the corresponding distribution coreset feature.
The ground truth used for training is a one-hot vector, where the distribution coreset index closest to the true center feature vector is one, and the cross-entropy loss is calculated with the MLP output.
To address the overconfidence of the trained deep neural network models, temperature scaling \cite{guo2017calibration} with temperature $T=2$ is applied to make the likelihood values more realistic.

Position information $\mathbf{x}$ is also crucial and can significantly affect the probability of $\Phi_i(\mathbf{x})$ being a normal feature, especially in object-type images.
To capture the position information, we generate $p(c_\mathrm{dist} | \mathbf{x})$ by accumulating the indices of $C_\mathrm{dist}$ for each position $\mathbf{x}$ in all training images $\forall x_i$, using Algorithm \ref{alg:pos_generate}.
In this process, features in the $p \times p$ neighborhood are accumulated in the histogram for robust estimation.
\begin{algorithm}[t]
	\caption{Calculation of $p(c_{\mathrm{dist}}|\mathbf{x})$}
    \label{alg:pos_generate}
	\begin{algorithmic}[1]
        \State Initialize $\mathrm{hist}(\mathord{\cdot}|\mathbf{x})$ as a zero vector of $\mathbb{R}^{|c_{\mathrm{dist}}|}$ for all $\mathbf{x}$
        \For {all training images $x_i$}
    		\For {all coordinates $\mathbf{x}$}
                \State $\mathrm{idx} \leftarrow$ find an index of nearest $c_{\mathrm{dist}}$ to $\Phi_i(\mathbf{x})$
                \State $\mathrm{hist}(\mathrm{idx}|\mathbf{x}) \leftarrow \mathrm{hist}(\mathrm{idx}|\mathbf{x}) + 1$
    		\EndFor
        \EndFor
        \State $p(c_{\mathrm{dist}}|\mathbf{x}) \leftarrow \mathrm{normalize} \left(\mathrm{hist} \left(\mathord{\cdot}|\mathbf{x} \right) \right)$
    \end{algorithmic} 
\end{algorithm}

To calculate $p(\Phi_i(\mathbf{x}) | c_\mathrm{emb}, \Omega)$ in \eqref{final_likelihood}, we assume that $p(\Phi_i(\mathbf{x}) | c_\mathrm{emb})$ is independent of $\Omega$, since $c_\mathrm{emb}$ contains all the information in $\Omega$ related to $\Phi_i(\mathbf{x})$.
Then, $p(\Phi_i(\mathbf{x}) | c_\mathrm{emb})$ is expressed in terms of an exponent of the distance between $\Phi_i(\mathbf{x})$ and $c_\mathrm{emb}$ as in most existing methods:
\begin{equation}
\label{approximate_condition_prob}
\begin {aligned}
p(\Phi_i(\mathbf{x}) | c_\mathrm{emb}, \Omega) \approx  p(\Phi_i(\mathbf{x}) | c_\mathrm{emb}) \approx e^{-\lambda {||\Phi_i(\mathbf{x}) - c_\mathrm{emb}||}_2},
\end {aligned}
\end{equation}
where $\lambda$ is a hyperparameter of an exponential function, and we set $\lambda = 1$ without optimizing.

\begin{figure}[t]
    \centering
    \includegraphics[width=\linewidth]{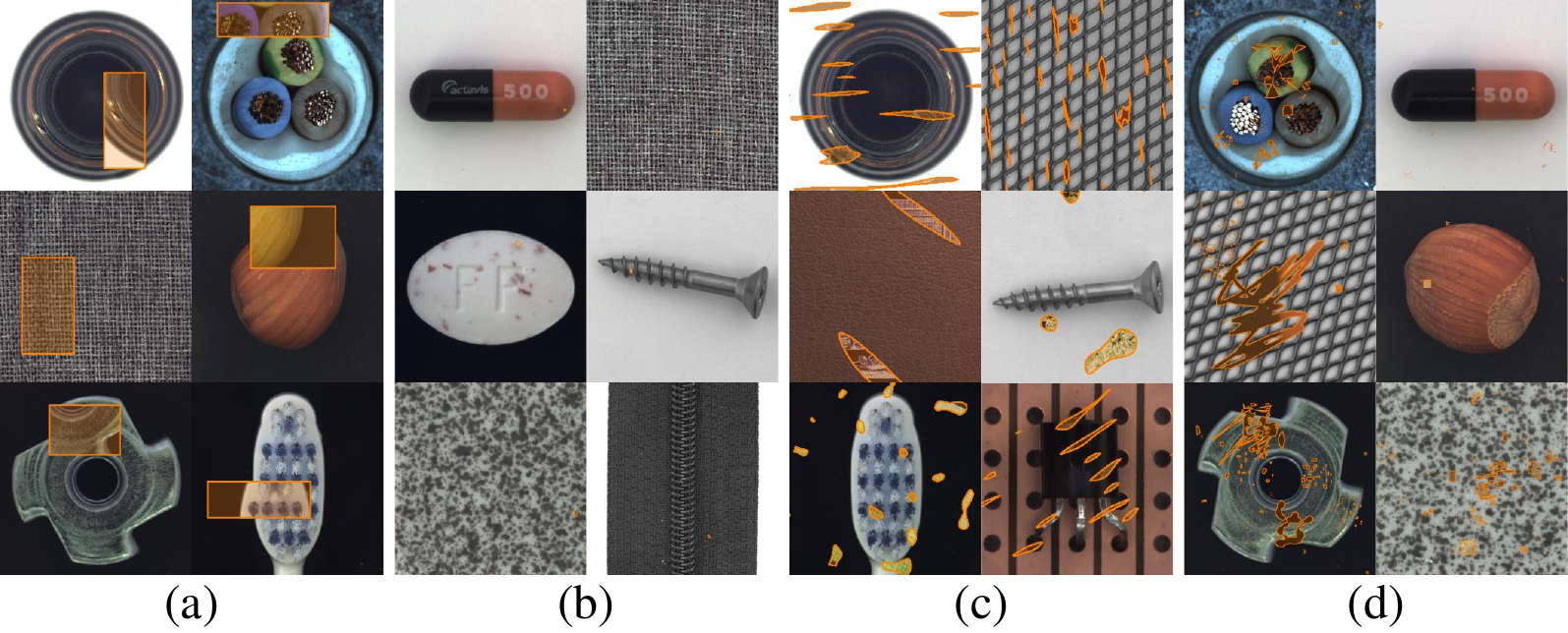}
    \vspace{-0.50cm}
    \caption
    {
       Examples of defect images generated by (a) CutPaste, (b) CutPaste (scar), (c) DR\AE M, and (d) manual drawing. The area corresponding to the defects are highlighted.
    }
    \label{fig:data_make}
    \vspace{-0.15cm}
\end{figure}

\subsection{Pixelwise Refinement}
We further improve the reliability of the anomaly map by using a refinement network $f$, trained in a supervised manner using an artificially created defect image dataset $\mathcal{D}$.
Let $\theta$ be parameters of $f$.
We aim to train optimal parameters
\begin{equation}
\theta^{*} = \arg \min_{\theta} \sum_{(I, \hat{A}, A) \in \mathcal{D}} \ell ( f(I, \hat{A}; \theta), A ).
\label{eq:problem_setting}
\end{equation}
$\mathcal{D}$ is composed of $(I, \hat{A}, A)$ pairs. 
$I$ is an artificially generated anomaly image, and $A$ represents the ground-truth anomaly map of $I$, with 1 assigned to defect regions and 0 assigned to others.
$\hat{A}$ is an anomaly map estimated from the proposed algorithm. We normalize each map into $[0,1]$. 
$\ell$ is a loss function between the refine $\Tilde{A} \triangleq f(I, \hat{A}; \theta)$ and the ground-truth $A$.

we create four different types of data for the dataset $\mathcal{D}$ with the same ratio. These four methods include CutPaste~\cite{li2021cutpaste}, CutPaste (scar)~\cite{li2021cutpaste}, DRÆM~\cite{zavrtanik2021draem}, and manual drawing.
As pointed out in CutPaste, training with defects of varying sizes and shapes together prevents the network from optimizing in a naive direction and enables better generalization performance. This is a significant advantage in cases where real abnormal data is unknown.
Figure~\ref{fig:data_make} shows the defect image examples generated by 4 methods from normal MVTec AD training data.
Defects generated by each method have distinct characteristics. CutPaste creates rectangular defects in larger areas, while CutPaste (scar) produces more detailed and thinner defects. DR\AE M and manual methods generate a more complex variety defect patterns.

\begin{figure}[t]
    \centering
    \includegraphics[width=\linewidth]{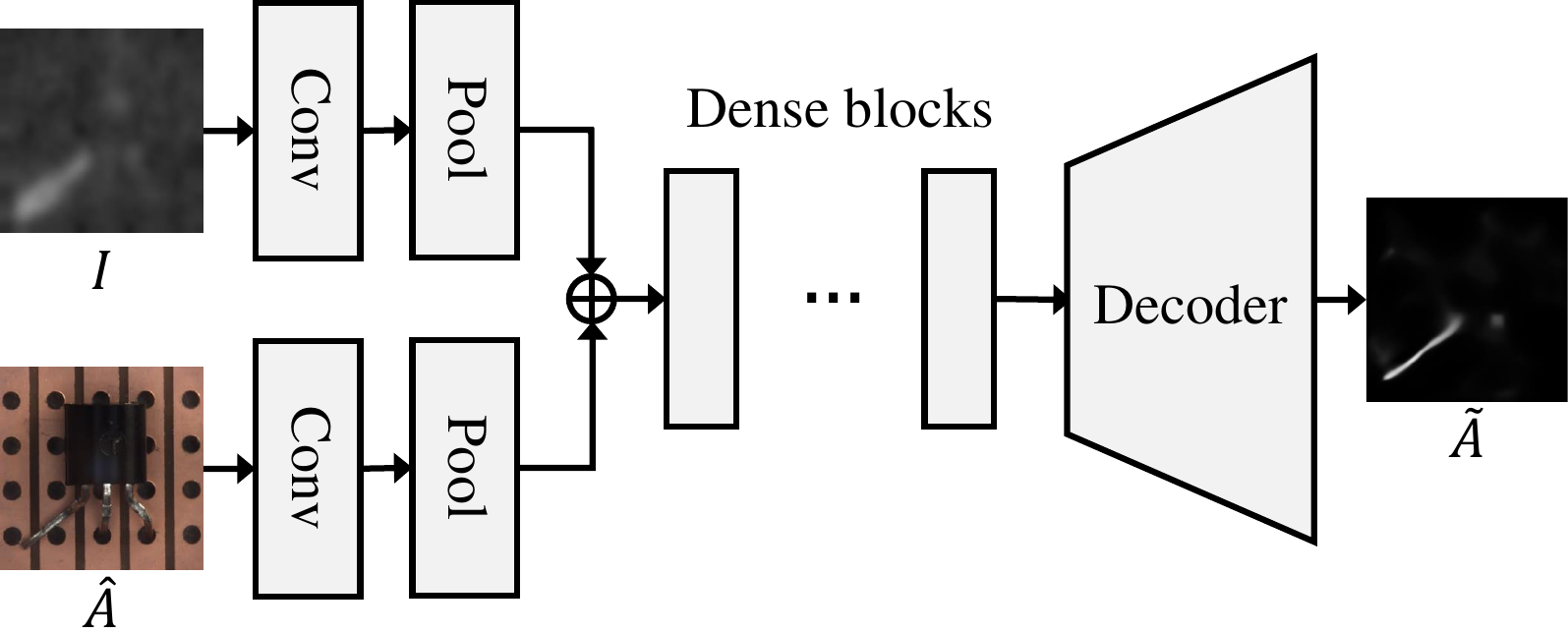}
    \vspace{-0.25cm}
    \caption
    {
       Schematic structure of the refinement network.
    }
    \label{fig:refinement_network}
\end{figure}

We adopt the encoder-decoder architecture for $f$. The network structure is based on \cite{lee2020multi} that uses DenseNet161~\cite{huang2017densely} as the backbone, but we introduce two modifications to it. First, the refinement network takes 4-channel inputs of an RGB image $I$ and an anomaly map $\hat{A}$. Second, we apply the early fusion method~\cite{imran2019depth} and fuse the features of $I$ and $\hat{A}$ after the first convolution layer. A schematic structure of the pixel-wise refinement network is presented in Figure~\ref{fig:refinement_network}.

To train $f$, we use a loss function consisting of two terms: $\ell = (\ell_\mathrm{reg} + \ell_\mathrm{grad}) / 2$.
The regression loss $\ell_\mathrm{reg}$ is calculated using L2-norm between $\Tilde{A}$ and $A$. 
\begin{equation}
\ell_\mathrm{reg} = \frac{|| \Tilde{A} - A ||_2}{HW},
\end{equation}
where $H$ and $W$ are the width and height of $A$.
Next, the gradient loss $\ell_\mathrm{grad}$ is 
\begin{equation}
\ell_\mathrm{grad} = \frac{|| \nabla_\mathrm{h} \Tilde{A} - \nabla_\mathrm{h} A ||_2 + || \nabla_\mathrm{w} \Tilde{A} - \nabla_\mathrm{w} A ||_2}{2HW}, 
\end{equation}
where $\nabla_\mathrm{h}$ and $\nabla_\mathrm{w}$ are partial derivative operations in the vertical and horizontal directions, respectively. $\ell_\mathrm{grad}$ improves the refinement results by making the network's training more concentrated near the edges of the defect region.

\begin{table*}[t]
    \scriptsize
    \addtolength{\tabcolsep}{1.2pt}
    \renewcommand{\arraystretch}{1.15}
    \caption{Anomaly detection and localization AUROC scores on MVTec AD \cite{bergmann2019mvtec} are presented. The first and second numbers indicate I-AUROC (image-level detection score) and P-AUROC (pixel-level localization score), respectively. Sub-total averages are provided for object and texture categories. For each category, the best result is \textbf{boldfaced}.}
    \vspace*{-0.2cm}
    \centering
    \begin{tabular}{cl | r@{\enspace}l r@{\enspace}l r@{\enspace}l r@{\enspace}l r@{\enspace}l r@{\enspace}l r@{\enspace}l | r@{\enspace}l}
    \toprule
    & & 
    \multicolumn{2}{c}{RIAD~\cite{zavrtanik2021reconstruction}} &
    \multicolumn{2}{c}{InTra~\cite{pirnay2022inpainting}} &
    \multicolumn{2}{c}{CutPaste~\cite{li2021cutpaste}} &
    \multicolumn{2}{c}{FastFlow~\cite{yu2021fastflow}} &
    \multicolumn{2}{c}{Tsai~\etal~\cite{tsai2022multi}} &
    \multicolumn{2}{c}{CFLOW-AD~\cite{gudovskiy2022cflow}} &
    \multicolumn{2}{c|}{PatchCore~\cite{roth2022towards}} &
    \multicolumn{2}{c}{\textbf{PNI}} \\
    \midrule
    \multirow{11}{*}{Object}

    & Bottle        & 99.9  & 98.4    & \textbf{100}   & 97.1    & 98.3  & 97.6    & \textbf{100}   & 97.7    & \textbf{100}   & 98.6    & \textbf{100}   & 98.76   & \textbf{100}   & 98.6    & \textbf{100}   & \textbf{98.87} \\
    & Cable         & 81.9  & 84.2    & 70.3  & 91.0    & 80.6  & 90.0    & \textbf{100}   & 98.4    & 98.8  & 98.2    & 97.59 & 97.64   & 99.5  & 98.4    & 99.76 & \textbf{99.10} \\
    & Capsule       & 88.4  & 92.8    & 86.5  & 97.7    & 96.2  & 97.4    & \textbf{100}   & 99.1    & 97.2  & 97.9    & 97.68 & 98.98   & 98.1  & 98.8    & 99.72 & \textbf{99.34} \\
    & Hazelnut      & 83.3  & 96.1    & 95.7  & 98.3    & 97.3  & 97.3    & \textbf{100}   & 99.1    & 99.6  & 97.8    & 99.98 & 98.82   & \textbf{100}   & 98.7    & \textbf{100}   & \textbf{99.37} \\
    & Metal nut     & 88.5  & 92.5    & 96.9  & 93.3    & 99.3  & 93.1    & \textbf{100}   & 98.5    & 97.8  & 99.1    & 99.26 & 98.56   & \textbf{100}   & 98.4    & \textbf{100}   & \textbf{99.29} \\
    & Pill          & 83.8  & 95.7    & 90.2  & 98.3    & 92.4  & 95.7    & \textbf{99.4}  & \textbf{99.2}    & 97.7  & 98.8    & 96.82 & 98.95   & 96.6  & 97.4    & 96.89 & 99.03 \\
    & Screw         & 84.5  & 98.8    & 95.7  & 99.5    & 86.3  & 96.7    & 97.8  & 99.4    & 94.1  & 98.5    & 91.89 & 98.10   & \textbf{98.1}  & 99.4    & 99.51 & \textbf{99.60} \\
    & Toothbrush    & \textbf{100}   & 98.9    & \textbf{100}   & 98.9    & 98.3  & 98.1    & 94.4  & 98.9    & \textbf{100}   & 99.0    & 99.65 & 98.56   & \textbf{100}   & 98.7    & 99.72 & \textbf{99.09} \\
    & Transistor    & 90.9  & 87.7    & 95.8  & 96.1    & 95.5  & 93.0    & 99.8  & 97.3    & 98.9  & 97.7    & 95.21 & 93.28   & \textbf{100}   & 96.3    & \textbf{100}   & \textbf{98.04} \\
    & Zipper        & 98.1  & 97.8    & 99.4  & 99.2    & 99.4  & 99.3    & 99.5  & 98.7    & 99.5  & 98.6    & 98.48 & 98.41   & 99.4  & 98.8    & \textbf{99.87} & \textbf{99.43} \\
    \cmidrule(lr){2-18}
    & Average       & 89.9  & 94.3    & 93.0  & 96.9    & 94.3  & 95.8    & 99.1 & 98.6   & 98.4  & 98.4    & 97.66 & 98.01   & 99.2 & 98.4   & \textbf{99.55} & \textbf{99.12} \\
    \cmidrule(lr){1-18}
    \multirow{6}{*}{Texture}
    & Carpet        & 84.2  & 96.3    & 98.8  & 99.2    & 93.1  & 98.3    & \textbf{100}   & \textbf{99.4}    & 93.4  & 98.4    & 98.73 & 99.23   & 98.7  & 99.0    & \textbf{100}   & \textbf{99.40} \\
    & Grid          & 99.6  & 98.8    & \textbf{100}   & 98.8    & 99.9  & 97.5    & 99.7  & 98.3    & \textbf{100}   & 98.5    & 99.60 & 96.89   & 98.2  & 98.7    & 98.41 & \textbf{99.20} \\
    & Leather       & \textbf{100}   & 99.4    & \textbf{100}   & 99.5    & \textbf{100}   & 99.5    & \textbf{100}   & 99.5    & 99.3  & 99.1    & \textbf{100}   & \textbf{99.61}   & \textbf{100}   & 99.3    & \textbf{100}   & 99.56 \\
    & Tile          & 98.7  & 89.1    & 98.2  & 94.4    & 93.4  & 90.5    & \textbf{100}   & 96.3    & 96.2  & 94.4    & 99.88 & 97.71   & 98.7  & 95.6    & \textbf{100}   & \textbf{98.40} \\
    & Wood          & 93.0  & 85.8    & 97.5  & 88.7    & 98.6  & 95.5    & \textbf{100}   & 97.0    & 99.7  & \textbf{97.5}    & 99.12 & 94.49   & 99.2  & 95.0    & 99.56 & 97.04 \\
    \cmidrule(lr){2-18}
    & Average       & 95.1  & 93.9    & 98.9  & 96.1    & 97.0  & 96.3    & \textbf{99.9} & 98.1    & 97.7  & 97.6    & 99.47 & 97.59   & 99.0 & 97.5   & 99.59 & \textbf{98.72} \\
    \cmidrule(lr){1-18}
    \multicolumn{2}{c|}{Average} 
                    & 91.7  & 94.2    & 95.0  & 96.6    & 95.2  & 96.0    & 99.4 & 98.5   & 98.1  & 98.1    & 98.26 & 97.87   & 99.1 & 98.1   & \textbf{99.56} & \textbf{98.98} \\
    \bottomrule

    \end{tabular}
    \label{tb:anomaly_performance_mvtec}
\end{table*}

\section{Experimental Results}
\subsection{Implementation Details}

\paragraph{Datasets}
We adopt two popular industrial datasets, MVTec AD \cite{bergmann2019mvtec} and BTAD \cite{mishra2021vt} to evaluate the proposed PNI. MVTec AD includes 15 subcategories, consisting of 10 object categories and 5 texture categories. The dataset contains a total of $5,354$ color images, including $3,629$ defect-free training images and $1,725$ test images that include both normal and anomalous images with ground-truth defect masks. Anomalous images are labeled with various types of defects.
BTAD is an industrial anomaly detection dataset with 3 subcategories and a total of $2,830$ color images. Of these, $1,800$ training images are normal, and the remaining test images include both normal and anomalous images with ground-truth masks.
As in \cite{cohen2020sub, defard2021padim, yi2020patch, roth2022towards}, images from all datasets are resized and center-cropped to remove negligible boundary pixels. 
Each image is resized to $512 \times 512$ and center-cropped to $480 \times 480$.

\vspace{-0.35cm}
\paragraph{Evaluation Metrics}
To access the performance of the proposed PNI, we use two metrics, AUROC (Area Under the Receiver Operator Curve) and AUPRO (per-region-overlap curve), as done in \cite{roth2022towards,tsai2022multi,deng2022anomaly}. 
AUROC is measured at the image level (I-AUROC) for anomaly detection performance and at the pixel level (P-AUROC) for anomaly localization performance.
AUPRO evaluates the anomaly localization performance by assigning equal weight to anomalous regions of different sizes in the image. AUPRO addresses the drawback of P-AUROC, where a prediction result in a single large anomalous region may have a greater impact than those in many small anomalous regions. High AUPRO indicates that the algorithm provides good anomaly localization results for both large and small anomalous regions.

\vspace{-0.35cm}
\paragraph{Parameter Setup}
Similar to~\cite{li2021cutpaste,roth2022towards}, we trained two models: a single network-based model and an ensemble network-based model. 
For the single model, WideResNet-101~\cite{zagoruyko2016wide} pre-trained on ImageNet~\cite{deng2009imagenet} data is used as the feature extractor. 
Also, for the ensemble model, ResNext-101 \cite{xie2017aggregated} and DenseNet-201~\cite{huang2017densely} are additionally used as feature extractors. 
The subsampling ratio to generate the embedding coreset is set to $0.01$, and the size of the distribution coreset $|C_\mathrm{dist}|$ is set to $2,048$. 
In the training process of $p(c_\mathrm{dist}|\mathbf{x})$ and $p(c_\mathrm{dist} |N_p(\mathbf{x}))$, the patch size of the neighborhood $p$ is set to 9.
The MLP network for the normal feature distribution consists of 10 fully-connected layers and each layer includes $2,048$ neurons. 
We train the MLP network using the Adam optimizer~\cite{kingma2014adam} for 15 epochs with a $10^{-3}$ learning rate and batch size $2,048$. 
Also, we adopt step learning rate decaying~\cite{he2019bag} with $\gamma = 0.1$ and 5 step size.
We don't use any data augmentation since each category has different permissible augmentation based on the characteristics of the images.

In the training process of the refinement network, we use the Adam optimizer for 60,000 iterations with a $10^{-4}$ learning rate and batch size 8. 
Also, we perform online data augmentation, including random horizontal flip, rotation, and color change in an online manner. In inference, we fuse the refined $\Tilde{A}$ at a 10\% ratio with $\hat{A}$ to obtain the final anomaly map. 

\subsection{MVTec AD}
\paragraph{AUROC}
Table~\ref{tb:anomaly_performance_mvtec} shows the performance of anomaly detection and localization for 15 subcategories of the MVTec AD dataset. We compare our proposed PNI algorithm with several conventional algorithms~\cite{zavrtanik2021reconstruction, pirnay2022inpainting, li2021cutpaste, yu2021fastflow, tsai2022multi, gudovskiy2022cflow, roth2022towards} in terms of I-AUROC and P-AUROC. Here, the results of single model versions of the proposed PNI algorithm, CutPaste, and PatchCore are compared. Also, the results of CFLOW-AD use an evaluation protocol that selects the best results from training with various hyperparameters. Some of the conventional algorithms provide multiple models by varying the experimental settings, and we provide detailed information on this in the supplementary document.

\begin{table}[!t]
    \scriptsize
    \setlength{\tabcolsep}{4.5pt}
    \renewcommand{\arraystretch}{1.15}
    \caption{Comparison of anomaly detection and localization results on MVTec AD~\cite{bergmann2019mvtec}. The proposed PNI is compared to recent algorithms in terms of I-AUROC, P-AUROC, and AUPRO. For AUPRO, sub-total averages are provided for both object and texture subcategories additionally.}
    \vspace*{-0.2cm}
    \centering
    \begin{tabular}{lllllll}
    \toprule
                & \multicolumn{2}{c}{AUROC} & & \multicolumn{3}{c}{AUPRO} \\
                \cmidrule(lr){2-3} \cmidrule(lr){5-7}
                & \multicolumn{1}{c}{Image} 
                & \multicolumn{1}{c}{Pixel} 
                & 
                & \multicolumn{1}{c}{Object} 
                & \multicolumn{1}{c}{Texture} 
                & \multicolumn{1}{c}{Average} \\
    \midrule
    Patch SVDD~\cite{yi2020patch}                       & 92.1  & 95.7  & & -     & -     & -     \\
    SPADE~\cite{cohen2020sub}                           & 85.5  & 96.0  & & 93.4  & 88.4  & 91.7  \\
    PaDiM~\cite{defard2021padim}                        & 95.3  & 97.5  & & 91.6  & 93.1  & 92.1  \\
    RIAD~\cite{zavrtanik2021reconstruction}             & 91.7  & 94.2  & & -     & -     & -     \\
    CutPaste~\cite{li2021cutpaste}                      & 95.2  & 96.0  & & -     & -     & -     \\
    DR\AE M~\cite{zavrtanik2021draem}                   & 98.0  & 97.3  & & -     & -     & -     \\
    FastFlow~\cite{yu2021fastflow}                      & 99.4  & 98.5  & & -     & -     & -     \\
    SOMAD~\cite{li2021anomaly}                          & 97.9  & 97.8  & & 94.1  & 91.6  & 93.3  \\
    InTra~\cite{pirnay2022inpainting}                   & 95.0  & 96.6  & & -     & -     & -     \\
    MB-PFM~\cite{wan2022unsupervised}                   & 97.5  & 97.3  & & 92.3  & 94.6  & 93.0  \\
    NSA~\cite{schluter2022natural}                      & 97.2  & 96.3  & & 90.4  & 92.2  & 91.0  \\
    IKD~\cite{cao2022informative}                       & -     & 97.81 & & 93.30 & 91.05 & 92.55 \\
    PatchCore~\cite{roth2022towards}                    & 99.1  & 98.1  & & 93.3  & 93.6  & 93.4  \\
    Reverse Distillation~\cite{deng2022anomaly}         & 98.5  & 97.8  & & 93.4  & 95.0  & 93.9  \\
    Tsai~\etal~\cite{tsai2022multi}                     & 98.1  & 98.1  & & 95.7  & 95.0  & 95.5  \\
    PEFM~\cite{wan2022position}                         & -     & 98.30 & & 95.30 & \textbf{95.95} & 95.52 \\
    CDO~\cite{cao2023collaborative}                     & -     & 98.22 & & 94.57 & 94.90 & 94.68 \\
    \textbf{PNI}                                   & \textbf{99.56} & \textbf{98.98} & & \textbf{96.34} & 95.47 & \textbf{96.05} \\
    \cmidrule(lr){1-7}
    Uniformed Students~\cite{bergmann2020uninformed}    & -     & -     & & 90.8  & 92.7  & 91.4  \\
    CutPaste (ensemble)~\cite{li2021cutpaste}           & 96.1  & -     & & -     & -     & -     \\
    PatchCore (ensemble)~\cite{roth2022towards}         & 99.6  & 98.2  & & -     & -     & 94.9  \\
    CFLOW-AD~\cite{gudovskiy2022cflow}                  & 98.26 & 98.62 & & 93.58 & 96.65 & 94.60 \\
    \textbf{PNI (Ensemble)}                        & \textbf{99.63} & \textbf{99.06} & & \textbf{96.83} & \textbf{96.00} & \textbf{96.55} \\
    \bottomrule
    \end{tabular}   
    \label{tb:mvtec_result}
    \vspace*{-0.2cm}
\end{table}

The proposed PNI algorithm shows the best anomaly detection performance of 99.56\% I-AUROC and anomaly localization performance of 98.98\% P-AUROC, surpassing FastFlow by 0.16\% and 0.48\%, respectively. Although FastFlow with CaiT shows high performance based on a powerful transformer, there is a noticeable performance drop in a few classes, leading to a decrease in the average score.
Our motivation for using position and neighborhood information can be more beneficial in object-type images, and the results support this guess. PNI significantly outperforms conventional algorithms in object classes, with considerable improvements in the localization of transistor, metal nut, and bottle classes. Moreover, the average I-AUROC of PNI for object classes is 99.55\%, showing a 43.8\% reduction in error compared to the second-best PatchCore.
Using neighborhood information for each pixel is also effective in measuring the anomaly at that position, even for texture classes. 
In addition, pixel-wise refinement works effectively on texture subcategories. The synthesized defect images used in network training handle various and complex defects occurring in real texture images well.

\begin{figure*}
\centering
\includegraphics[width=\textwidth]{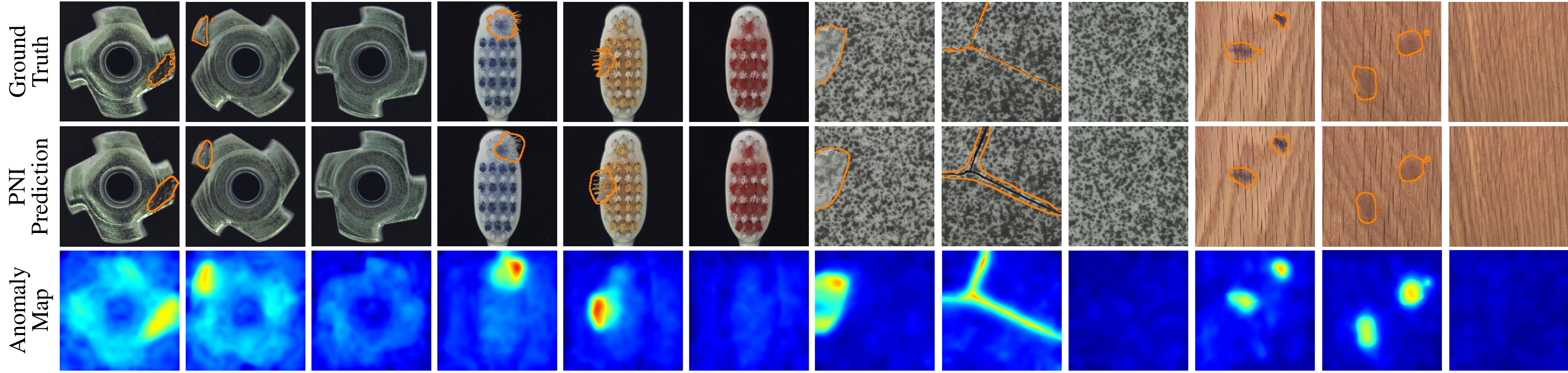}
\caption{Visualization of anomaly localization results of PNI on the MVTec AD. Input images with ground-truth masks (top), predicted masks (mid) and predicted anomaly maps (bottom) are provided. }
\label{example_mask}
\vspace{-0.30cm}
\end{figure*}

\vspace{-0.10cm}
\paragraph{AUPRO}
Table~\ref{tb:mvtec_result} compares the PNI with conventional algorithms, including the AUPRO metric. The performance of each algorithm is compared in terms of the average for object categories, texture categories, and overall. Detailed performance for all subcategories is discussed in the supplemental document. Some algorithms~\cite{bergmann2020uninformed, li2021cutpaste, roth2022towards, gudovskiy2022cflow} propose models that use multiple networks. At the bottom of Table~\ref{tb:mvtec_result}, we compare those results with the ensemble network-based PNI.  

PNI also shows superior performance over conventional algorithms in AUPRO. In overall AUPRO, PNI outperforms the second-best PFFM~\cite{wan2022position} by 0.53\%, with a score of 96.05\%. As mentioned in the previous analysis, our approach of using position and neighborhood information is more effective for object subcategories, and PNI surpasses the second-best Tsai~\etal~\cite{tsai2022multi} by 0.64\%. In the texture subcategories as well, PNI shows the second-best performance.
Furthermore, PNI (Ensemble) exhibits the best performance in AUROC and AUPRO, object, and texture subcategories, surpassing the results of all conventional algorithms without exception.

\begin{table}[!t]
    \scriptsize
    \setlength{\tabcolsep}{2.1pt}
    \renewcommand{\arraystretch}{1.15}
    \caption{The Ablation Results on MVTec AD. Anomaly detection and localization performance are measured in I-AUROC [\%] and P-AUROC [\%], respectively.}
    \vspace*{-0.2cm}
    \centering
    \begin{tabular}{ccc cccccc}
    \toprule
                                \multirow{2}{*}{Neighbor} & \multirow{2}{*}{Position} & \multirow{2}{*}{Refine \ } 
                                & \multicolumn{3}{c}{I-AUROC} & \multicolumn{3}{c}{P-AUROC} \\
                                \cmidrule(lr){4-6} \cmidrule(lr){7-9}
                                & & & Object & Texture & Average & Object & Texture & Average \\
    \midrule
                                 - & - & -                  & 99.01 & 98.75 & 98.92     & 98.70 & 97.15 & 98.18 \\
                                \cmark & - & -              & 99.38 & 99.55 & 99.44     & 98.79 & 98.29 & 98.62 \\
                                \cmark & \cmark & -         & 99.46 & 99.46 & 99.46     & 99.04 & 98.33 & 98.80 \\
                                \cmark & \cmark & \cmark    & \textbf{99.55} & \textbf{99.59} & \textbf{99.56}    & \textbf{99.12} & \textbf{98.72} & \textbf{98.98} \\
    \bottomrule
    \end{tabular}
    \label{tb:ablation_result}
    \vspace*{-0.2cm}
\end{table}

\vspace{-0.10cm}
\paragraph{Ablation Study}
We conducted an ablation study to verify the effect of using the three components of our proposed PNI algorithm: neighborhood information, position information, and pixel-wise refinement. Table~\ref{tb:ablation_result} shows the results, and the following observations can be made:
\begin{itemize}
\itemsep0mm
\item The baseline without the three components is identical to PatchCore and performs similarly.
    \vspace*{-0.05cm}
\item Since the baseline deals with normal features unconditionally, the remaining models in the ablation study outperform the performance of the baseline.
    \vspace*{-0.05cm}
\item The use of neighborhood information improves overall performance significantly, enhancing the anomaly detection performance from 98.92\% to 99.44\%, which reduces the error by approximately 48.1\%.
    \vspace*{-0.05cm}
\item Position information provides an additional gain in object subcategories, but there is little improvement observed in texture subcategories. This result makes sense as the motivation for using position information is irrelevant to texture subcategories.
    \vspace*{-0.05cm}
\item Pixel-wise refinement is more effective in texture subcategories and has complementary properties with position information.
\end{itemize}

\begin{figure}[t]
\centering
\subfloat{%
  \includegraphics[width=\columnwidth]{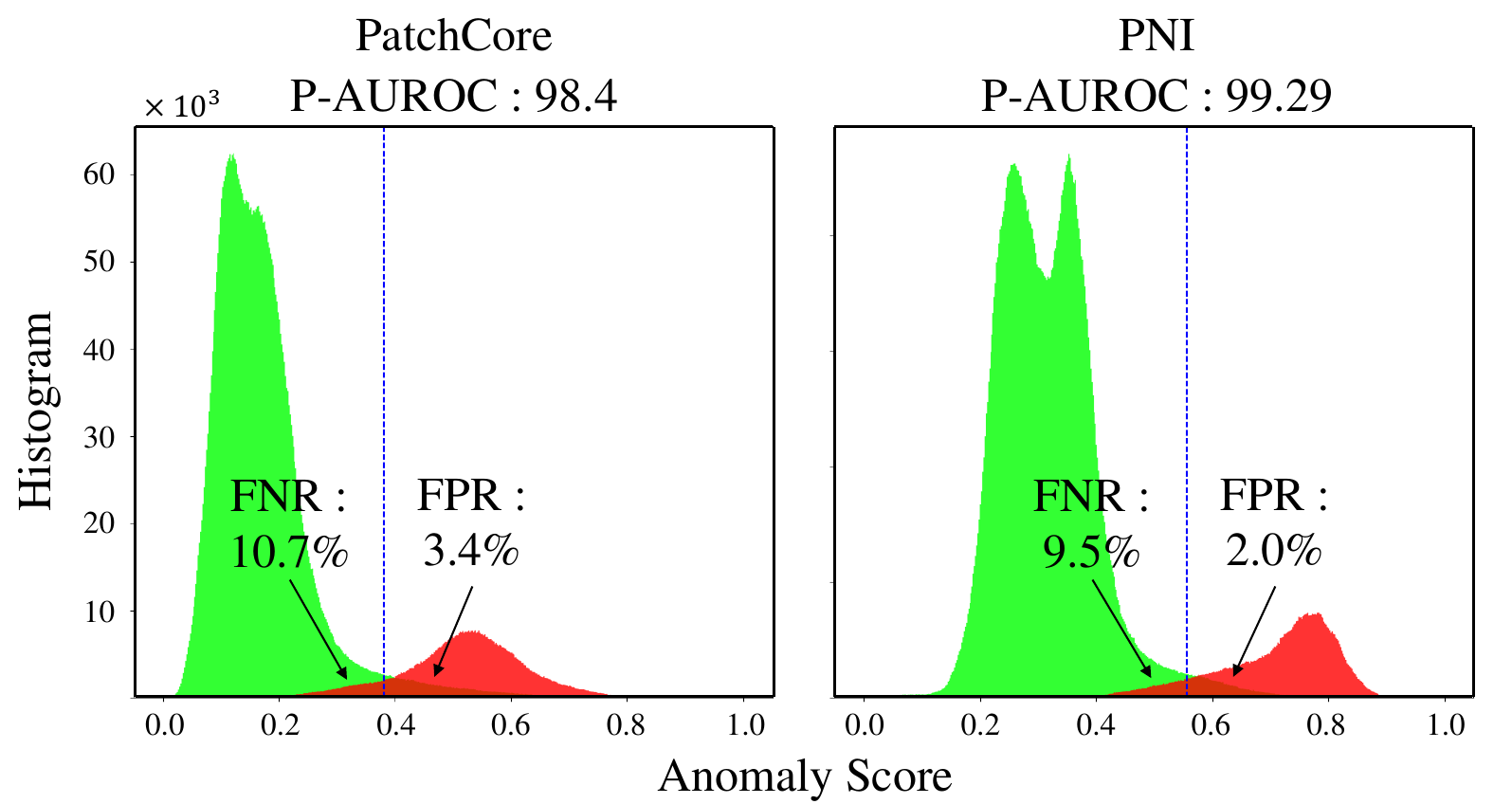}%
}
\caption{Histograms of anomaly scores for the metal nut subcategory in MVTec AD evaluated by PatchCore (left) and PNI (right) are shown. The green and red indicate the distribution of anomaly scores for normal and anomalous pixels, respectively. The blue vertical line indicates the threshold that maximizes the F1 score.}
\label{histogram_feature}
\vspace{-0.40cm}
\end{figure}

\paragraph{Qualitative Results}
Figure~\ref{example_mask} shows the results of anomaly localization performed by the proposed PNI algorithm on the test images of MVTec AD. The first row shows the ground-truth defect mask overlaid on the input image, and the second row shows the prediction results of PNI.
We generate the prediction mask using the threshold that maximizes the F1 score. 
The last row visualizes the anomaly map predicted by PNI.
The red parts in each image indicate high anomaly while the blue ones indicate low anomaly.
Qualitative results show that the predicted mask generally follows the ground truth, leading to performance improvement.

\paragraph{Analysis}
Figure \ref{histogram_feature} compares the histograms of PatchCore and the proposed PNI algorithm for pixel-wise anomaly scores in the metal nut subcategory. The green and red areas represent the distributions of anomaly scores for normal and abnormal pixels, respectively. Additionally, the blue vertical line indicates the threshold that optimizes the F1 score. The red area on the left of the threshold represents misclassified anomaly pixels or false negative pixels, while the green area on the right of the threshold represents false positive pixels. The FPR and FNR of PatchCore are 3.4\% and 10.7\%, respectively, which decrease to 2.0\% and 9.5\% in the PNI algorithm. These results correspond to the higher P-AUROC score of the proposed method.

Additionally, we computed image misclassification, false-positive and false-negative samples with the threshold optimizing F1 scores of anomaly detection.
Out of the 467 normal test images and 1258 defective test images, a total of 7 false-positive and 12 false-negative detection errors were found, which is a significant improvement compared to 19 false-positive and 23 false-negative errors of PatchCore.
We have provided detailed information on this in the supplementary document.

\subsection{BTAD}
The anomaly localization performance of the proposed PNI on the BTAD dataset is shown in Table \ref{tb:btad_result}.
We compare the anomaly localization performance of PNI with conventional algorithms~\cite{mishra2021vt,yi2020patch,yu2021fastflow,tsai2022multi}.
As shown in Table \ref{tb:btad_result}, the proposed model outperforms other state-of-the-art algorithms in anomaly localization on all product categories in BTAD as well as the average score.

\begin{table}[!t]
    \scriptsize
    \setlength{\tabcolsep}{2pt}
    \renewcommand{\arraystretch}{1.15}
    \caption{Anomaly localization results on BTAD~\cite{mishra2021vt} as measured in P-AUROC [\%].}
    \vspace*{-0.2cm}
    \centering
    \begin{tabular}{c|cccc|c}
    \toprule
    Products    & VT-ADL~\cite{mishra2021vt} & P-SVDD~\cite{yi2020patch} & FastFlow~\cite{yu2021fastflow} & Tsai~\etal~\cite{tsai2022multi} & \textbf{PNI} \\
    \midrule
    1           & 76.3 & 94.9 & 95 & 97.3 & \textbf{97.4}\\
    2           & 88.9 & 92.7 & 96 & 96.8 & \textbf{97.0}\\
    3           & 80.3 & 91.7 & \textbf{99} & \textbf{99.0} & \textbf{99.0}\\
    \cmidrule(lr){1-6}
    Average     & 81.8 & 93.1 & 97 & 97.7 & \textbf{97.8}\\
    \bottomrule
    \end{tabular}
    \label{tb:btad_result}
    \vspace*{-0.2cm}
\end{table}

\vspace{0.20cm}
\section{Conclusion}
We propose a new algorithm, PNI, for industrial anomaly detection and localization that accurately estimates the distribution of normal features by incorporating position and neighborhood information. PNI models position information using accumulated histograms from normal training images and uses a multi-layer perceptron network to model the normal feature distribution given neighborhood information. Additionally, PNI introduces a pixel-wise refinement network using synthesized anomaly images to improve the anomaly map according to the input image, which is the first refinement approach in the field of industrial anomaly detection and localization as far as the authors know. Various experiments demonstrate the overall performance and effectiveness of the proposed PNI algorithm.

\clearpage

\clearpage

{\small
\bibliographystyle{ieee_fullname}
\bibliography{egbib}
}

\end{document}